\documentclass{article}

    \usepackage[preprint]{neurips_2022}

\usepackage[utf8]{inputenc} % allow utf-8 input
\usepackage[T1]{fontenc}    % use 8-bit T1 fonts
\usepackage{hyperref}       % hyperlinks
\usepackage{url}            % simple URL typesetting
\usepackage{booktabs}       % professional-quality tables
\usepackage{amsfonts}       % blackboard math symbols
\usepackage{nicefrac}       % compact symbols for 1/2, etc.
\usepackage{microtype}      % microtypography
\usepackage{xcolor}         % colors
\usepackage{graphicx}

\usepackage{amsthm}
\usepackage{MnSymbol}
\usepackage{amsmath}
\hypersetup{
    colorlinks=true,
    linkcolor=blue,
    citecolor=black,
    urlcolor=blue,
}
\usepackage{float}
\title{A survey on Self Supervised learning approaches for improving Multimodal representation learning}

\author{%
  Naman Goyal \\
  Department of Computer Science\\
  Columbia University\\
  \texttt{ng2848@columbia.edu} \\
}

\begin{document}

\maketitle

\begin{abstract}
    Recently self supervised learning has seen explosive growth and use in variety of machine learning tasks because of its ability to avoid the cost of
annotating large-scale datasets.
  This paper gives an overview for best self supervised learning approaches for multimodal learning. The presented approaches have been aggregated by extensive study of the literature and tackle the application of self supervised learning in different ways. The approaches discussed are cross modal generation, cross modal pretraining, cyclic translation, and generating unimodal labels in self supervised fashion.
\end{abstract}

% \begin{figure}
%   \centering
%   \fbox{\rule[-.5cm]{0cm}{4cm} \rule[-.5cm]{4cm}{0cm}}
%   \caption{Sample figure caption.}
% \end{figure}
% \begin{table}
%   \caption{Sample table title}
%   \label{sample-table}
%   \centering
%   \begin{tabular}{lll}
%     \toprule
%     \multicolumn{2}{c}{Part}                   \\
%     \cmidrule(r){1-2}
%     Name     & Description     & Size ($\mu$m) \\
%     \midrule
%     Dendrite & Input terminal  & $\sim$100     \\
%     Axon     & Output terminal & $\sim$10      \\
%     Soma     & Cell body       & up to $10^6$  \\
%     \bottomrule
%   \end{tabular}
% \end{table}

\section{Introduction}

\begin{figure}
\centering
\includegraphics[width=0.9\textwidth]{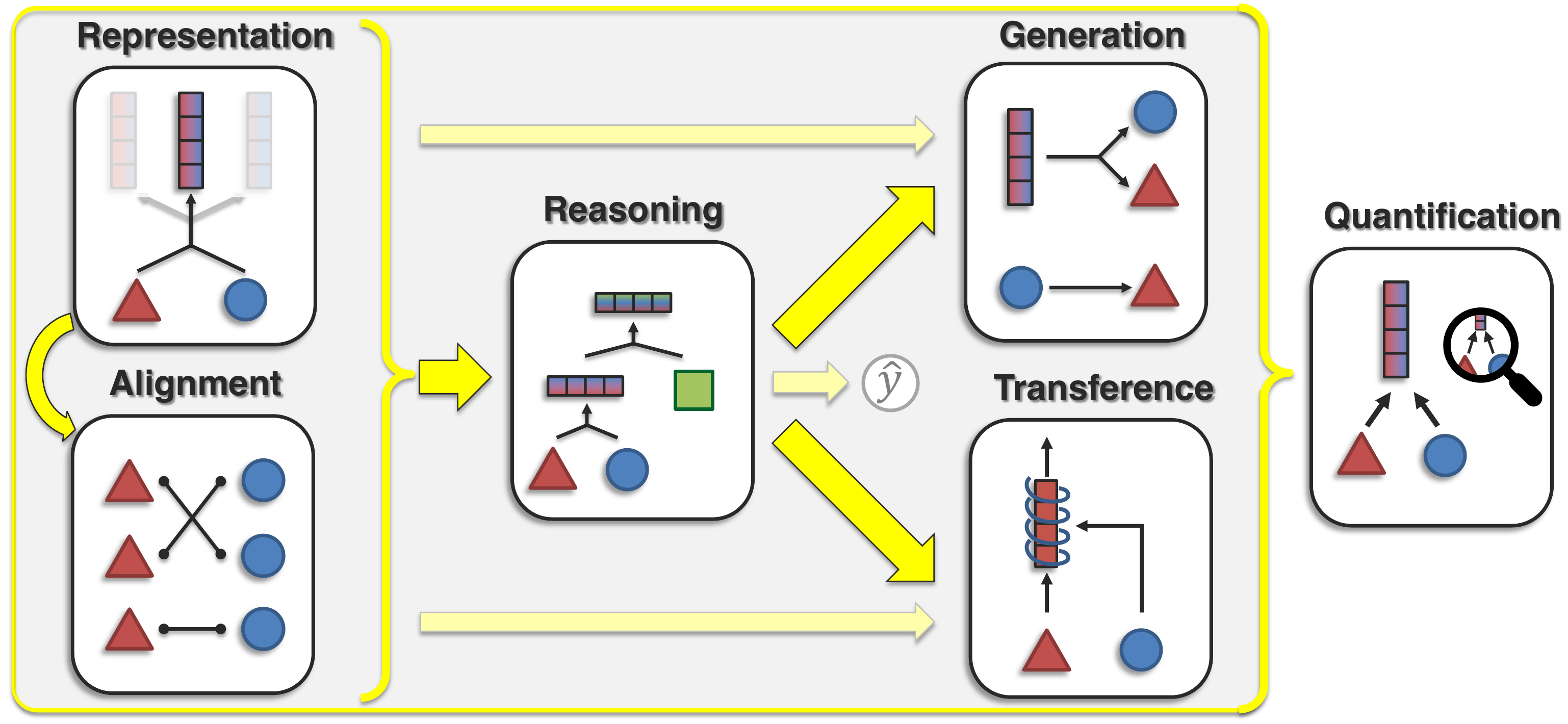}
 \caption{Taxonomy of various challenges in the multimodal learning. A triangle and circle represents elements from two different modalities. (1) \textit{Representation} studies how to represent and summarize multimodal
data to reflect the heterogeneity and interconnections between individual modality elements. (2) \textit{Alignment} aims to identify the
connections and interactions across all elements. (3) \textit{Reasoning} aims to compose knowledge from multimodal evidence usually
through multiple inferential steps for a task. (4) \textit{Generation} involves learning a generative process to produce raw modalities that
reflect cross-modal interactions, structure, and coherence. (5) \textit{Transference} aims to transfer knowledge between modalities and their
representations. (6) \textit{Quantification} involves empirical and theoretical studies to better understand heterogeneity, interconnections,
and the multimodal learning process.}
 \label{fig:challenges}
\end{figure}

Multimodal machine learning is a vibrant multi-disciplinary research field that aims to design intelligent systems for understanding, reasoning, and learning through integrating multiple communicative modalities, including linguistic, acoustic, visual, tactile, and physiological messages.  Multimodal learning attracts intensive research interest
because of broad applications such as intelligent tutoring \citep{petrovica2017emotion},
robotics \citep{noda2014multimodal}, and healthcare \citep{frantzidis2010classification}. Generally speaking, existing research efforts mainly focus on how to fuse multimodal data effectively and how to learn a good representation for each modality. Further the extensive survey by \cite{liang2022foundations} gives a taxonomy of the challenges in multimodal learning as showing in Figure \ref{fig:challenges}.

Self-supervised learning \citep{jaiswal2020survey} obtains supervisory signals from the data itself, often leveraging the underlying structure in the data. The general technique of self-supervised learning is to predict any unobserved or hidden part (or property) of the input from any observed or unhidden part of the input. For example, as is common in NLP, we can hide part of a sentence and predict the hidden words from the remaining words. 

The aim of this paper is to provide an overview of best self supervised learning approaches which tackle the representation chellenge in multimodal learning. We present 4 approaches after extensive survey of literature in the domain.

\begin{enumerate}
    \item We look at cross modal generation which basically for a given image-text pair, generates image-to-text and text-to-image. We then compare the generated text and image samples with the input pair.
    \item We then look at cross modal transformer which uses cues from different modality namely audio and video to do predict the masked token in masked language modelling.
    \item We then look at the approach of cyclic translation between modalities using a Seq2Seq network. A given modality is translated to another modality and then back translated. The learned hidden encoding is used for final prediction.
    \item Finally we look at approach of generating unimodal labels from multimodal datasets in self supervised fashion. Multitask learning is used to jointly train on both multimodal and unimodal labels.
\end{enumerate}

For brevity purpose we give a high level detail of each method while omit the superficial details and refer the reader to the original work for further reference. 

\section{Methodology}

\subsection{Cross-modal generation}

\begin{figure}
\centering
\includegraphics[width=0.7\textwidth]{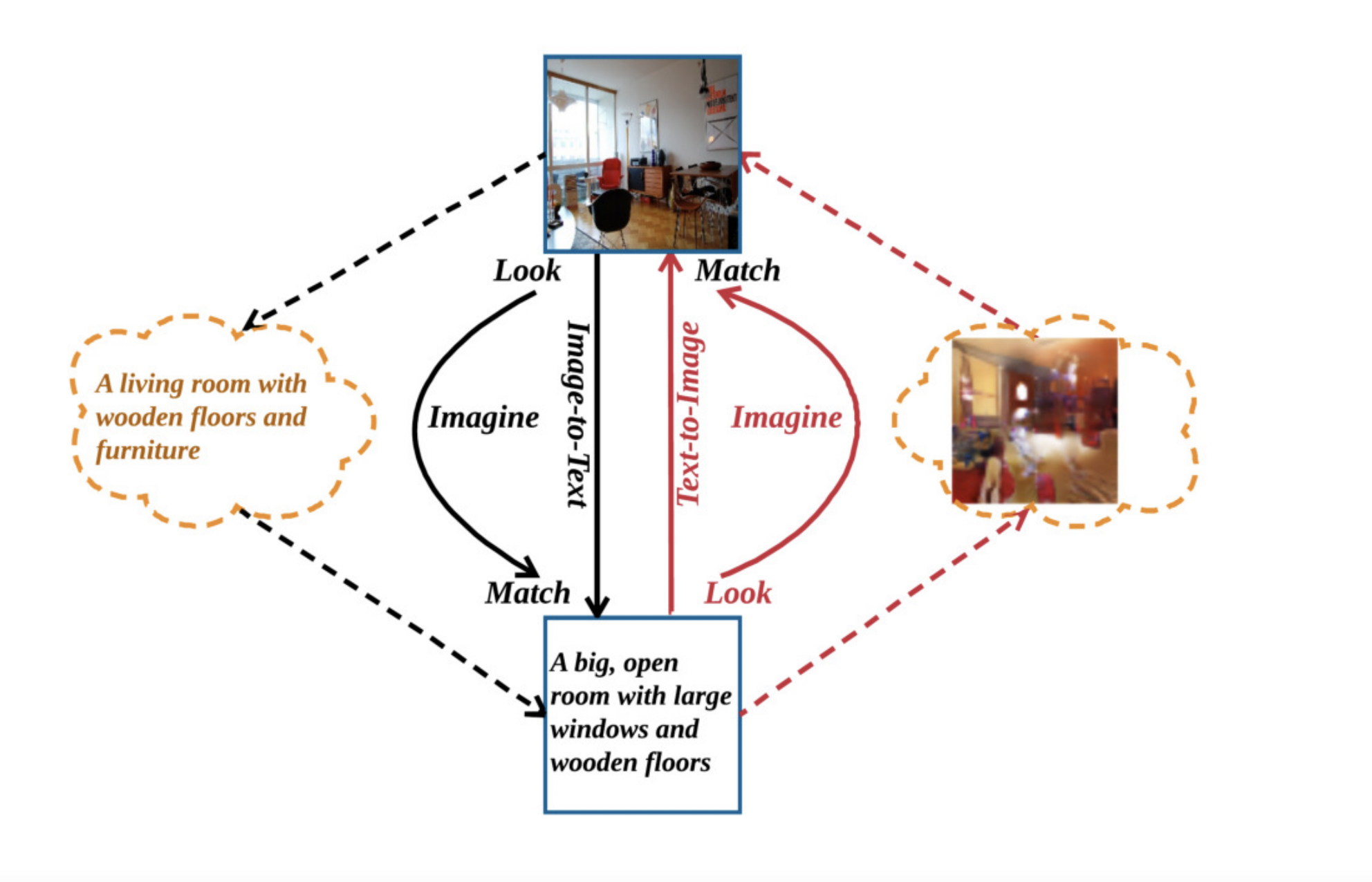}
 \caption{Conceptual illustration of the crossmodal feature embedding with generative models. The cross-modal retrievals (Image-to-Text and Text-to-Image) are shown in different color. The two blue boxes are crossmodal data, and the generated data are shown in two dashed
yellow clouds.}
 \label{fig:cross1}
\end{figure}

The main idea proposed by \cite{gu2018look} is addition to  conventional cross-modal feature embedding at
the global semantic level, to introduce an additional
cross-modal feature embedding at the local level, which
is grounded by two generative models: image-to-text and
text-to-image. Figure \ref{fig:cross1} illustrates the concept of the proposed cross-modal feature embedding with generative models at high level, which includes three learning steps: \textit{look},
\textit{imagine}, and \textit{match}. 
Given a query in image or text,
first \textit{look} at the query to extract an abstract representation.
Then, \textit{imagine} what the target item (text or image) in
the other modality should look like, and get a more concrete grounded representation. This is accomplish by using the representation of one modality (to be estimated) to
generate the item in the other modality, and comparing the
generated items with gold standards. After that, in \textit{match} step
the right image-text pairs using the relevance score which is
calculated based on a combination of grounded and abstract
representations.

\begin{figure}[H]
\centering
\includegraphics[width=\textwidth]{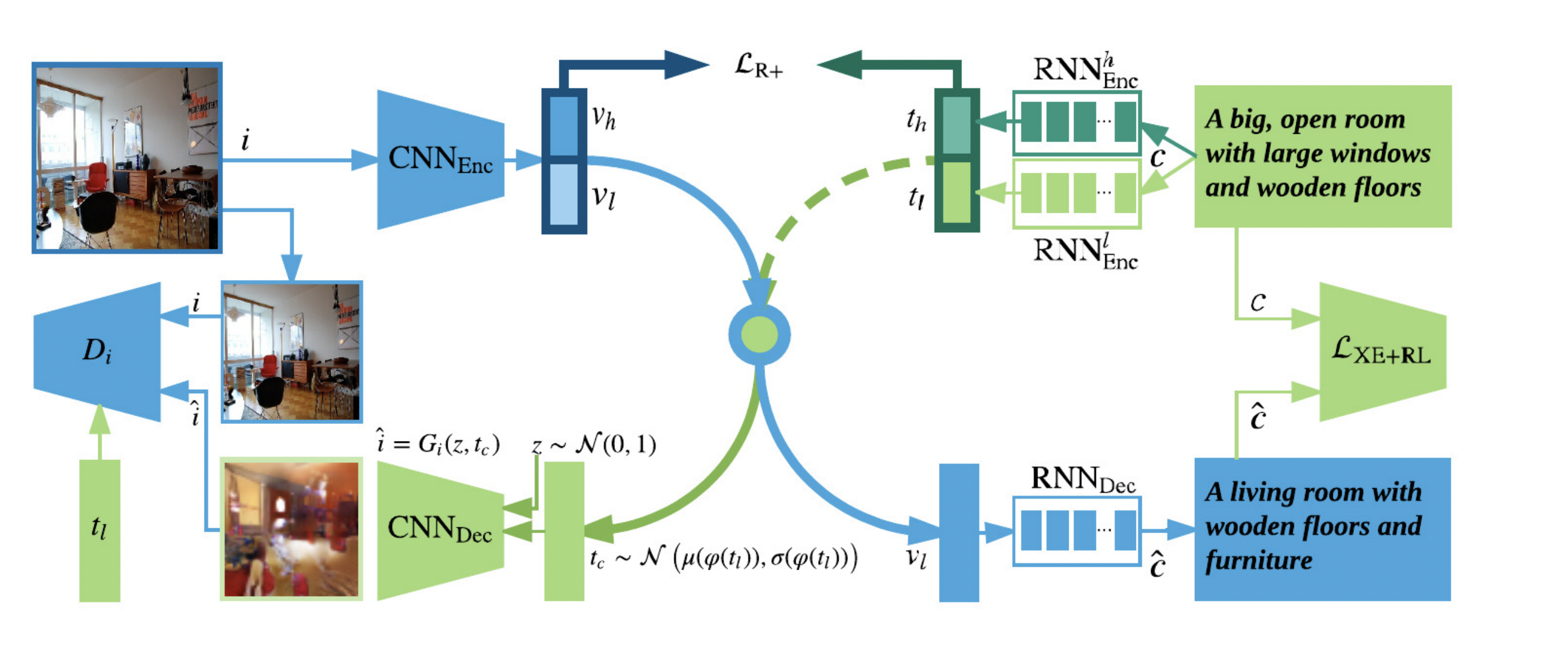}
 \caption{The generative cross-modal learning framework. The entire framework consists of three training
paths: cross-modal feature embedding (the entire upper part), image-to-text generative feature learning (the blue path), and text-to-image generative adversarial feature learning (the green path). It includes six networks: two sentence encoders RNN$^{h}_{Enc}$ (dark green) and RNN$^{l}_{Enc}$ (light green), one image encoder CNN$_{Enc}$ (blue), one sentence  decoder RNN$_{Dec}$, one image decoder CNN$_{Dec}$ and one discriminator $D_{i}$ .}
 \label{fig:cross2}
\end{figure}

\paragraph{Architecture} Figure \ref{fig:cross2} shows the overall architecture for the proposed
generative cross-modal feature learning framework. The entire system consists of three training paths:
multi-modal feature embedding (the entire upper part),
image-to-text generative feature learning (the blue path),
and text-to-image generative adversarial feature learning
(the green path). The first path is similar to the existing
cross-modal feature embedding that maps different modality features into a common space. However, the difference here is that they use two branches of feature embedding, i.e.,
making the embedded visual feature $v_h$ (resp. $v_l$) and the textual feature $t_h$ (resp. $t_l$) closer. They consider ($v_h$, $t_h$) as high-level abstract features and ($v_l$ , $t_l$) as detailed grounded features. The grounded features will be used and regularized in the other two generative feature learning paths. The entire first training path mainly includes one image encode CNN$_{Enc}$ and two sentence encoders RNN$^h_{Enc}$ and RNN$^l_{Enc}$. The second training path (the blue path) is to generate a sentence from the embedded generative visual feature $v_l$.It consists of the image encoder CNN$_{Enc}$ and a sentence detector RNN$_{Dec}$. With a proper loss against ground-truth sentences, the grounded feature $v_l$ will be adjusted via back propagation. The third training path (the green path) is to generate an image from the textual feature $t_l$. Here we adopt the generative adversarial model, which comprises a generator / decoder CNN$_{Dec}$ and a discriminator $D_i$.

Overall, through these two paths of cross-modal generative feature learning, authors hope to learn powerful cross-modal
feature representations. During the testing stage, {$v_h$, $v_l$}
and {$t_h$, $t_l$} will be used as the final feature representations
for cross-modal retrieval.

\subsection{Cross-modal pretraining for transformers}

The main idea proposed by \cite{khare2021self} is to pretrain a cross-modal transformer in self supervised fashion and then finetune the transformer for emotion recognition. This allows for attending to specific input features (visual frames, words, speech segments) that are relevant to the task. 

The first part of the model architecture achieves this by using self-attention based transformer encoder for individual modalities. After encoding from unimodal encoder, authors obtain the self-attended outputs $S_{\mathbb{A}}$, $S_{\mathbb{V}}$ and $S_{\mathbb{T}}$ from the audio, visual and text modalities respectively.

\begin{figure}[H]
\centering
\includegraphics[width=\textwidth]{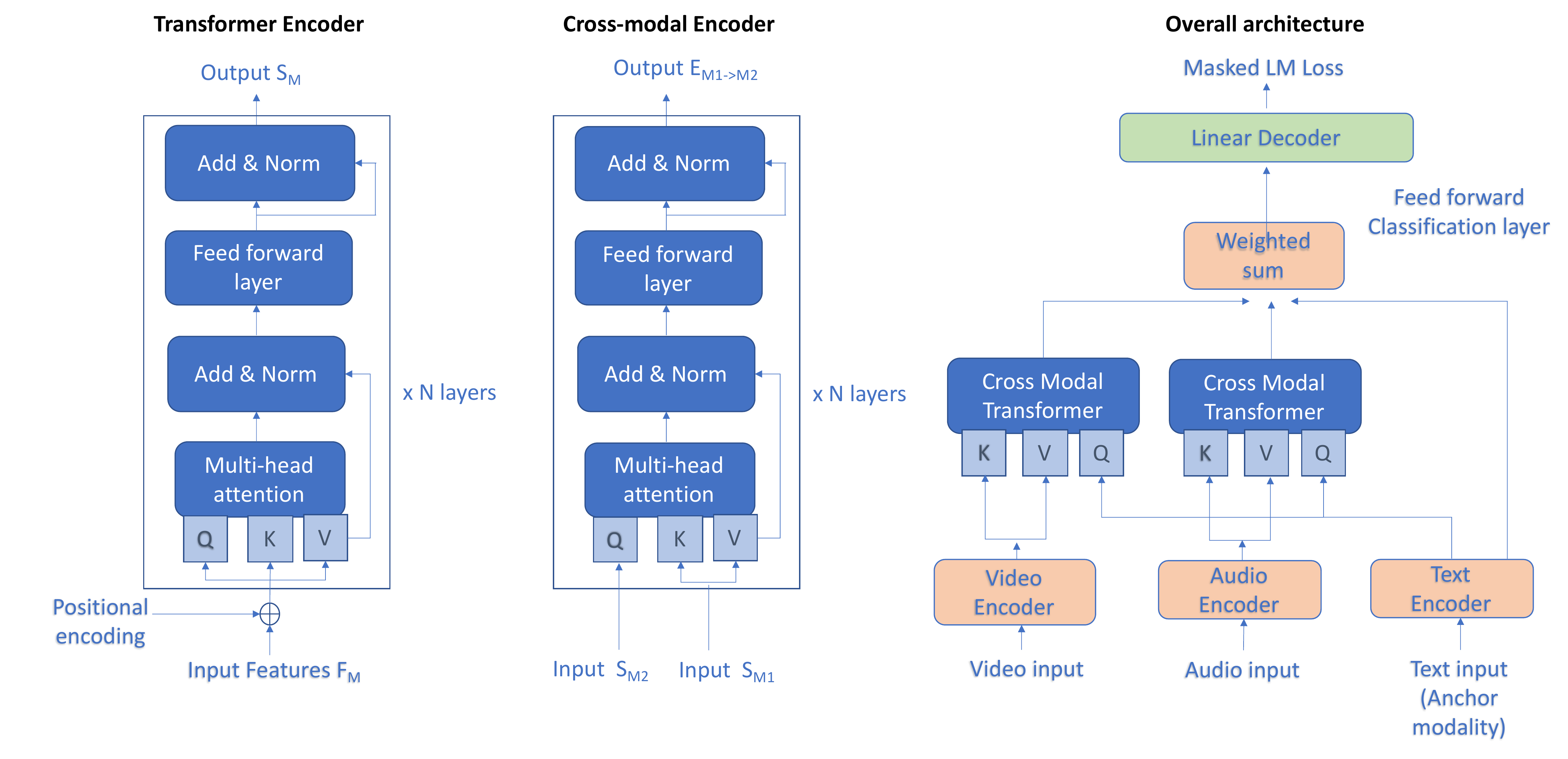}
 \caption{Cross-modal transformer based self-supervised learning architecture. The figure shows the self-attention based transformer encoder layers, the cross-modal attention encoder module and the overall architecture of the self-supervised model.}
 \label{fig:crosstrans1}
\end{figure}

Next, the authors combine the uni-modal transformer encoder outputs, $S_{\mathbb{A}}$, $S_{\mathbb{V}}$ and $S_{\mathbb{T}}$, to learn the final multi-modal representation for emotion recognition. This is done by the cross-modal transformer, which computes the attention map between features from two different modalities $\mathbb{M}_1$ and $\mathbb{M}_2$. The cross-modal attention allows for increasing attention weights on features that are deemed important for emotion recognition by more than one modality.

\paragraph{Pretraining} For pretraining the model in self supervised fashion, the authors use a variation of masked LM task to train the model. They propose to predict words by looking at audio and visual context in addition to the context words around the masked input.\ Intuitively, the auxiliary information present in visual expressions and audio features like intonation would provide relevant input for predicting the masked words.\ For example, consider the phrase ``This movie is [MASK]'' as an input to the model. The [MASK] word could be predicted as ``amazing'' if the audio and visual features show that speaker is happy.\ Alternatively, the prediction could be ``terrible'' if the speaker seems discontent while talking. This information cannot be derived from text only input.\ They posit that the latent representations learned using the masked LM task with multi-modal input will not just encode context, but also the relevant emotional information which could be used to predict those words.

\subsection{Self supervised cyclic translation}

With the recent success of sequence to sequence (Seq2Seq) models in machine translation, there is an opportunity to explore new ways of learning joint representations that may not require all input modalities at test time. \cite{pham2019found} propose a method to learn robust joint representations by translating between modalities. Our method is based on the key insight that translation from a source to a target modality provides a method of learning joint representations using only the source modality as input. They augment modality translations with a cycle consistency loss to ensure that the joint representations retain maximal information from all modalities. Once the translation model is trained with paired multimodal data, one only needs data from the source modality at test time for final sentiment prediction.

\begin{figure}[H]
\centering
\includegraphics[width=.6\linewidth]{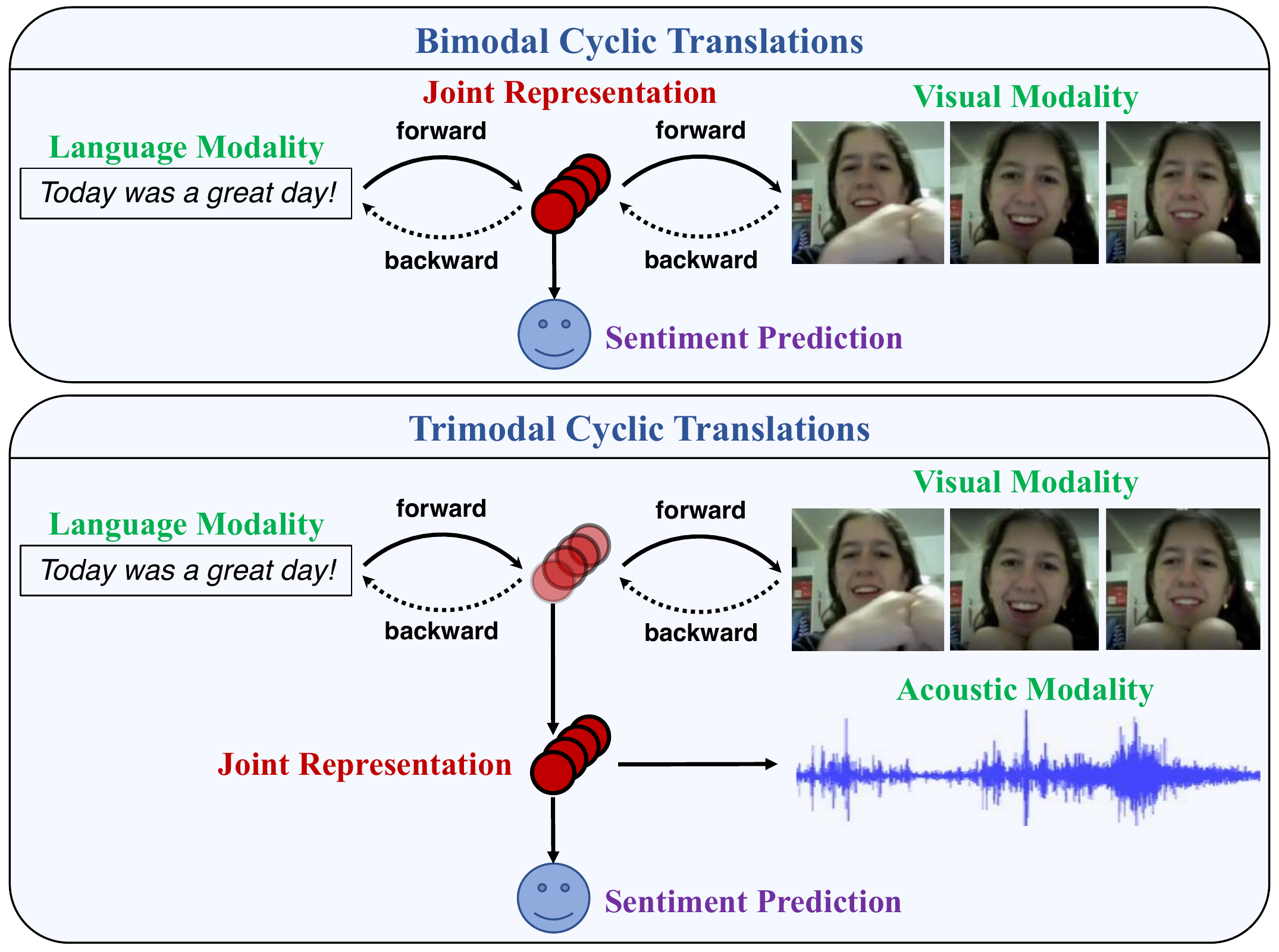}
\caption{
\small
{Learning robust joint representations via multimodal cyclic translations. Top: cyclic translations from a source modality (language) to a target modality (visual). Bottom: the representation learned between language and vision are further translated into the acoustic modality, forming the final joint representation. In both cases, the joint representation is then used for sentiment prediction.}}
\label{fig:cycle1}
\end{figure}

The proposed model learns robust joint multimodal representations by translating between modalities. Figure~\ref{fig:cycle1} illustrates these translations between two or three modalities. The method is based on the key insight that translation from a source modality $S$ to a target modality $T$ results in an intermediate representation that captures joint information between modalities $S$ and $T$. Author then extend this insight using a cyclic translation loss involving both \textit{forward translations} from source to target modalities, and \textit{backward translations} from the predicted target back to the source modality. Together, these \textit{multimodal cyclic translations}  ensure that the learned joint representations capture maximal information from both modalities. They also propose a hierarchical setting to learn joint representations between a source modality and multiple target modalities. The proposed approach is trainable end-to-end with a coupled translation-prediction loss which consists of (1) the cyclic translation loss, and (2) a prediction loss to ensure that the learned joint representations are task-specific (\textit{i.e.} multimodal sentiment analysis).

\begin{figure}[H]
\centering
\includegraphics[width=0.7\textwidth]{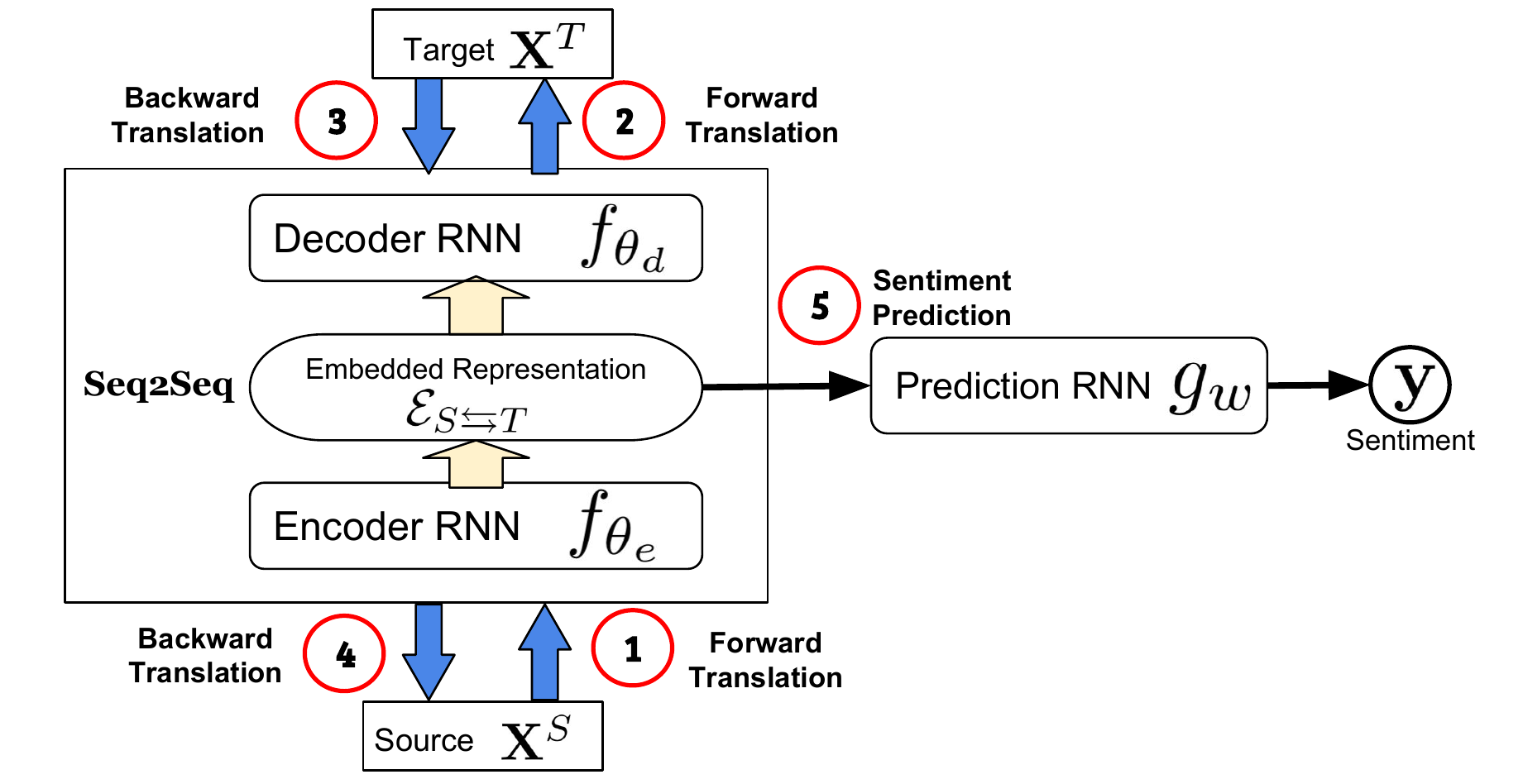}
\caption{
\small
{
The architecture for two modalities: the source modality $\mathbf{X}^{S}$ and the target modality $\mathbf{X}^{T}$. The joint representation ${\mathcal{E}}_{S \leftrightarrows T}$ is obtained via a cyclic translation between $\mathbf{X}^{S}$ and $\mathbf{X}^{T}$. Next, the joint representation ${\mathcal{E}}_{S \leftrightarrows T}$ is used for sentiment prediction. The model is trained end-to-end with a coupled translation-prediction objective. At test time, only the source modality $\mathbf{X}^{S}$ is required.}}
\label{fig:cycle2}
\end{figure}

\subsection{Self supervised unimodal label prediction and Multi-task for better representation}

The main idea prosposed by \cite{yu2021learning} is the that representation learning only using multimodal labels is only targets at learning differential information. But usually unimodal labels for a multimodal dataset are not available. Hence the authors propose a strategy to generate unimodal labels in self supervised fashion. These unimodal labels are joined used with multimodal labels to train the model. Further, during the training stage, authros design a weight-adjustment strategy to balance the learning progress among different subtasks.
\begin{figure}[H]
\centering
\includegraphics[width=0.8\textwidth]{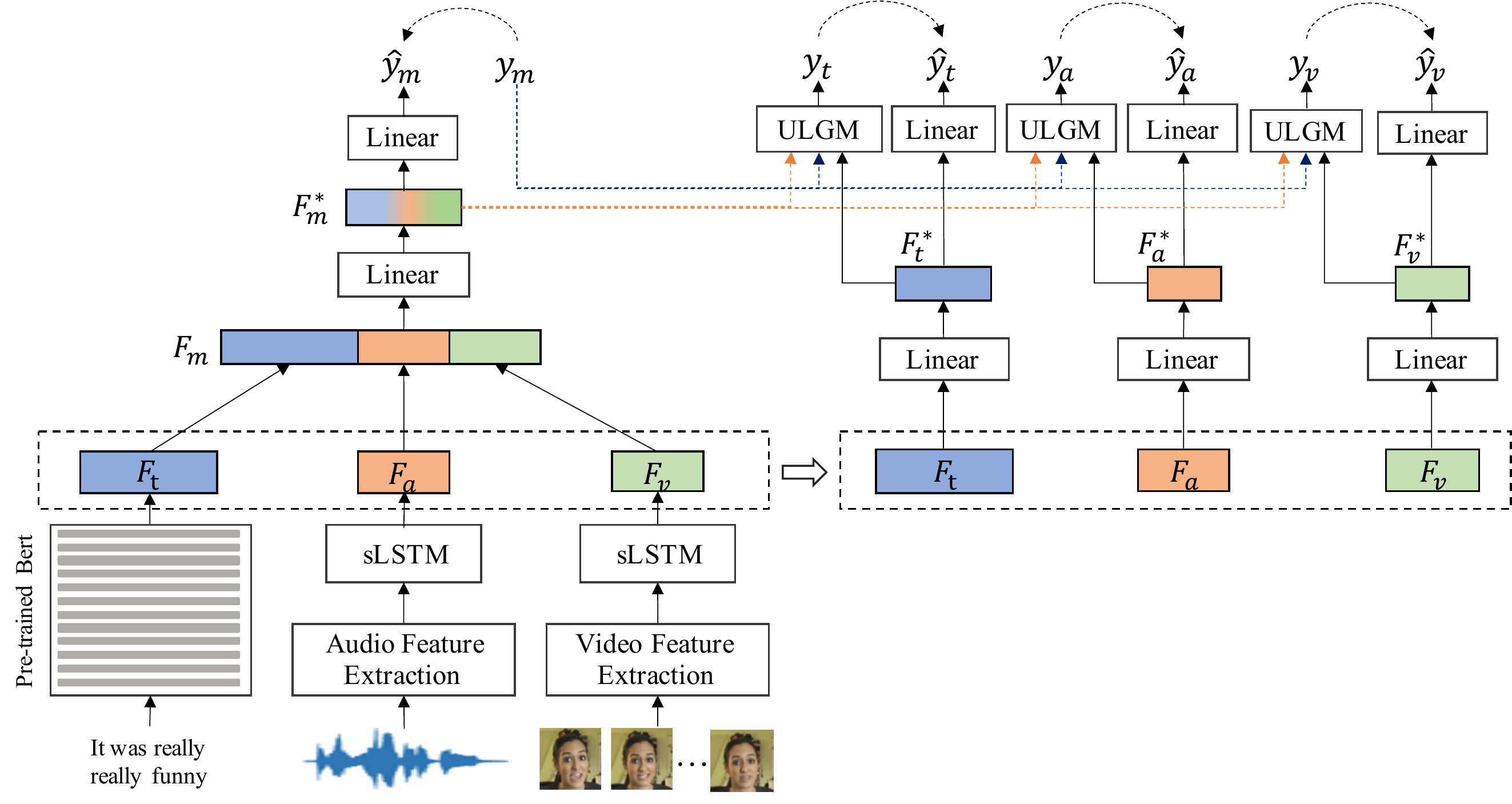}
\caption{The overall architecture of Self-Supervised Multi-task Multimodal model. 
    The $\hat{y}_m$, $\hat{y}_t$, $\hat{y}_a$, and $\hat{y}_v$ are 
    the predictive outputs of the multimodal task and the three unimodal tasks, respectively. 
    The $y_m$ is the multimodal annotation by human. 
    The $y_t$, $y_a$, and $y_v$ are the unimodal supervision 
    generated by the self-supervised strategy. Finally, $\hat{y}_m$ is used 
    as the sentiment output.}
\label{fig:uni1}
\end{figure}

\begin{figure}[H]
\centering
\includegraphics[width=0.5\textwidth]{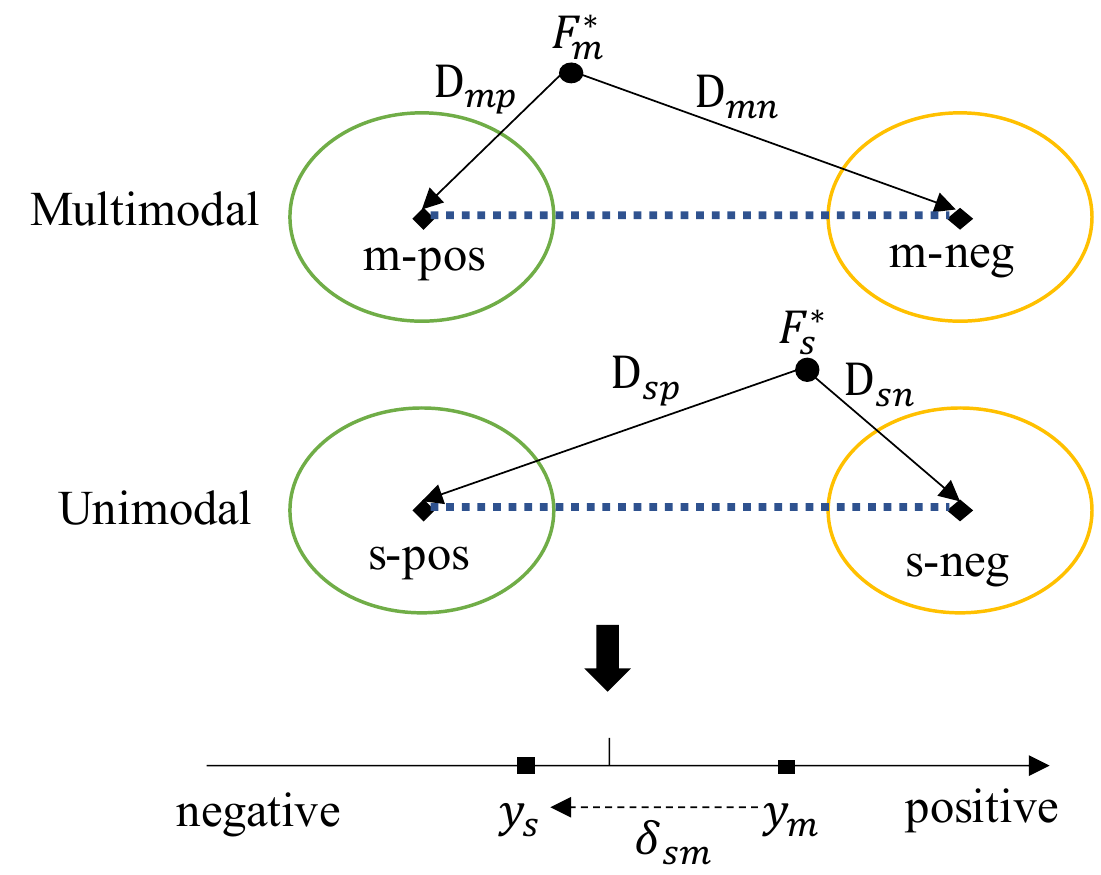}
\caption{Unimodal label generation. Multimodal representation $F_m^*$ is closer to the positive center (m-pos) while 
    unimodal representation is closer to the negative center (s-neg). 
    Therefore, unimodal supervision $y_s$ is 
    added a negative offset $\delta_{sm}$ to the multimodal label $y_m$ $\Rightarrow y_s = y_m + \alpha_s - \alpha_m$}
\label{fig:uni2}
\end{figure}

\paragraph{Unimodal Label Generation Module}
The ULGM aims to generate uni-modal supervision values based on  multimodal annotations and modality representations. In order to avoid  unnecessary interference with the update of network parameters, the ULGM is designed as a non-parameter module. Generally, unimodal supervision values are highly correlated  with multimodal labels. Therefore, the ULGM calculates the offset according to the relative distance from modality representations to class centers. It basically estimates the unimodal label from the multimodal label using the relative distance from the modality specific class center for the specific modality as shown in figure \ref{fig:uni2}.

\section{Conclusion}

This paper presents the best approaches to improve the learned representation in multimodal learning in a self supervised fashion. The various presented approaches are generic and work by different principles - cross modal generation, cross modal pretraining, cyclic translation, and generating unimodal learning. The field of self supervised learning is ever evolving and its application to other areas in machine learning are yet to fully realized.

\bibliographystyle{plainnat}
\bibliography{myrefs.bib}

%%%%%%%%%%%%%%%%%%%%%%%%%%%%%%%%%%%%%%%%%%%%%%%%%%%%%%%%%%%%

% \appendix

% \section{Appendix}

% Optionally include extra information (complete proofs, additional experiments and plots) in the appendix.
% This section will often be part of the supplemental material.

\end{document}